\documentclass[conference]{IEEEtran}

\usepackage{cite}
\usepackage{amsmath,amssymb,amsfonts}
\usepackage{graphicx}
\usepackage{textcomp}
\usepackage{xcolor}
\usepackage{booktabs}
\usepackage{tabularx}
\usepackage{array}
\usepackage{url}
\usepackage[T1]{fontenc}
\usepackage[hidelinks]{hyperref}

\newcolumntype{Y}{>{\centering\arraybackslash}X}

\begin{document}

\title{ASR-Style Transcript Normalization Removes Useful Pragmatic Signal for Dialogue Act Classification}

\author{
\IEEEauthorblockN{Lucky Onyekwelu-Udoka}
\IEEEauthorblockA{\textit{Department of Electrical and Computer Engineering} \\
\textit{Iowa State University}\\
Ames, Iowa, United States \\
lucky@iastate.edu}
\and
\IEEEauthorblockN{Md Shafiqul Islam}
\IEEEauthorblockA{\textit{Department of Electrical and Computer Engineering} \\
\textit{Iowa State University}\\
Ames, Iowa, United States \\
shafiqul@iastate.edu}
}

\maketitle

\begin{abstract}
Automatic speech recognition (ASR) pipelines often
normalize transcripts by removing filled pauses and
punctuation and expanding colloquial forms, potentially
discarding cues useful for dialogue act classification.
We evaluate nine transcript conditions across six
classifiers (TF-IDF logistic regression, RoBERTa, BERT,
ALBERT, GPT-2, and BiLSTM) and three conversational
corpora (Switchboard, ICSI MRDA, and SLUE-HVB),
supplemented by multi-seed experiments, normalization
controls, and real-ASR evaluations. Across five random
seeds, heavy normalization reduces Macro-F1 in all
15 neural model--corpus combinations under the original
preprocessing protocols. Removal using the original
filler and backchannel lexicon consistently degrades
performance on MRDA and SLUE-HVB but produces mixed
effects on Switchboard. Punctuation-aware TF-IDF
experiments show that punctuation removal reduces
Macro-F1 by 0.036 on Switchboard and 0.078 on MRDA,
demonstrating that the original linear baseline's
apparent insensitivity resulted from tokenization.
Restricting filler removal to \textit{um}, \textit{uh},
and \textit{erm} preserves degradation on MRDA but
largely eliminates the SLUE-HVB effect, where the
original lexicon removes class-defining backchannel
expressions. These findings demonstrate that
normalization effects depend on preprocessing
definitions, feature extraction, and conversational
context, supporting task-dependent rather than
indiscriminate transcript normalization.
\end{abstract}


\section{Introduction}

ASR transcripts are commonly normalized by dropping filled pauses, removing punctuation, and standardizing colloquial forms. These changes improve readability, but they can also erase signals that help manage turns and organize discourse \cite{CLARK200273, https://doi.org/10.1111/j.1749-818X.2008.00068.x, CLARK1989259}. Although disfluencies and recognition errors have been studied, how normalization itself affects dialogue act classification across models and interaction types is still not well characterized.

We investigate nine transcript conditions across
six classifiers (TF-IDF logistic regression, RoBERTa,
BERT, ALBERT, GPT-2, and BiLSTM) and three corpora
(Switchboard, ICSI MRDA, and SLUE-HVB), yielding
18 model--corpus combinations. The main experiments transform reference transcripts,
while separate analyses examine recognition errors
in real ASR output. We also assess tokenization,
lexicon sensitivity, and random-seed variability.

Our main contributions are:
\begin{enumerate}
    \item A multi-model, cross-corpus evaluation
    showing consistent Macro-F1 degradation under
    heavy normalization across five random seeds,
    with more variable effects from filler removal.

    \item A punctuation-aware TF-IDF analysis
    demonstrating that the apparent insensitivity
    of linear classifiers to punctuation removal
    results from tokenization rather than model
    architecture alone.

    \item A restricted filler-lexicon analysis
    showing that degradation persists on MRDA
    but largely disappears on SLUE-HVB when
    class-defining backchannel tokens are preserved.

    \item Cluster-level statistical analyses,
    real-ASR evaluations, and selective
    normalization experiments assessing the
    robustness and practical implications
    of transcript preprocessing.
\end{enumerate}

Our findings show that normalization effects depend
on the linguistic cues removed, feature extraction,
and conversational setting, supporting task-dependent
rather than indiscriminate transcript preprocessing.

\section{Related Work}

Disfluencies, filled pauses, and backchannels
convey information about hesitation, turn management,
and discourse structure
\cite{CLARK200273,
https://doi.org/10.1111/j.1749-818X.2008.00068.x,
shriberg1999phoneticCO,
GinzburgFernandezSchlangen2014,
CLARK1989259}. Removing these expressions during
transcript normalization may therefore eliminate
cues useful for downstream pragmatic inference.

Prior studies have investigated ASR recognition errors
and configurable transcription-noise evaluation
\cite{Cui2021AnAT,Nechaev2021,LI2024104598,
kubis-etal-2023-back,shapira-etal-2025-measuring,
zusag24_interspeech}.
However, recognition errors differ from deliberate
transcript normalization, which can alter linguistic
cues even when transcription is accurate.

Dialogue act recognition research has examined
lexical and conversational cues
\cite{jurafsky-shriberg-biasca1997,
Core1997CodingDW,stolcke-etal-2000-dialogue}.
Zelasko et al.\ \cite{zelasko-etal-2021-helps}
investigated the role of punctuation in
transformer-based dialogue act recognition and
segmentation. Prior research has compared lightweight and transformer-based
architectures for speech emotion classification through
systematic ablation studies \cite{onyekwelu2025emotion}.

Disfluency studies establish the communicative role
of hesitations and backchannels, whereas ASR-robustness
work examines recognition errors and Zelasko et al.\
focus on punctuation in dialogue-act recognition.
These studies do not jointly isolate deliberate
filler/backchannel removal, punctuation stripping,
and colloquial expansion for dialogue-act classification
across model families and conversational registers.
Our controlled text ablations address this gap with
nine conditions, six classifiers, and three corpora,
supplemented by tokenizer and lexicon controls,
multi-seed evaluation, and real-ASR experiments.

\section{Experimental Setup}

\textbf{Datasets.}
We evaluate three conversational corpora: Switchboard
\cite{chang-et-al2020}, ICSI MRDA
\cite{shriberg-et-al2004-mrda,wylupek-mrda},
and SLUE-HVB \cite{shon-et-al2023-slue2},
representing telephone conversations, multi-party
meetings, and scripted call-center dialogues,
respectively. SLUE-HVB is multi-label; approximately
40\% of utterances have multiple tags, and we use the
release's first tag as the primary class. Table~\ref{tab:datasets} summarizes
the datasets after removing labels with fewer than
50 instances in combined train--test counts. Switchboard uses a conversation-level
80/20 split, while MRDA and HVB use their
release-defined splits. The MRDA release describes its
canonical partition as meeting-disjoint, although the
reformatted records do not expose meeting identifiers.

\begin{table}[t]
\caption{Dataset statistics after rare-label filtering
($\geq 50$ occurrences).}
\label{tab:datasets}
\centering
\scriptsize
\setlength{\tabcolsep}{3pt}
\renewcommand{\arraystretch}{0.9}
\begin{tabularx}{0.95\columnwidth}{@{}lYYY@{}}
\toprule
& \textbf{Switchboard} & \textbf{MRDA} & \textbf{HVB} \\
\midrule
Register & Telephone & Meetings & Call-center \\
Train/test & 34.6k/9.1k & 75.1k/16.7k & 11.3k/6.1k \\
Labels & 27 & 12 & 16 \\
Canonical filler rate$^{a}$ & 40.5\% & 15.9\% & 5.7\% \\
\bottomrule
\end{tabularx}
\vspace{1pt}\parbox{0.95\columnwidth}{\tiny $^{a}$Utterances containing \textit{um}, \textit{uh}, or \textit{erm}; the Switchboard rate is 17,703/43,763 utterances.}
\end{table}

\textbf{Models.}
We evaluate TF-IDF with logistic regression
(TF-IDF+LR), RoBERTa-base \cite{liu-et-al2019-roberta},
BERT-base-cased \cite{devlin-et-al2018-bert},
ALBERT-base-v2 \cite{lan-et-al2019-albert},
GPT-2 \cite{radford-et-al2019-gpt2}, and BiLSTM
\cite{hochreiter-schmidhuber1997-lstm}.
TF-IDF+LR uses unigram/bigram features
($\mathrm{min\_df}=3$, $\mathrm{max\_df}=0.95$)
and class-balanced logistic regression.
Transformers are fine-tuned for three epochs
(learning rate $2\times10^{-5}$, batch size 16,
maximum length 96). BiLSTM uses 200-dimensional
embeddings, 128 hidden units, and 15 training
epochs. Neural experiments use five seeds while TF-IDF+LR is deterministic.
We additionally evaluate a punctuation-aware,
case-preserving TF-IDF tokenizer to examine
punctuation discarded by the default tokenizer.

\textbf{Normalization Conditions.}
We evaluate nine conditions:
\texttt{gold\_text} (original transcripts),
\texttt{norm\_colloq\_only} (colloquial expansion),
\texttt{rm\_punct\_only} (punctuation removal),
\texttt{rm\_discourse\_only} (removal of
\textit{like}/\textit{well}),
\texttt{gold\_alpha\_text} (corpus-specific alphanumeric form),
\texttt{rm\_fillers\_only} (filler/backchannel lexicon removal),
\texttt{clean\_light} (lowercasing and removal
of non-alphanumeric characters except apostrophes),
\texttt{clean\_heavy} (light cleaning, filler
removal, and colloquial expansion), and
\texttt{asr\_noise} (synthetic recognition errors).

Synthetic ASR noise applies token deletion (8\%),
substitution (8\%), and insertion (4\%).
Each main-grid condition is applied to both
training and test transcripts.

The original filler lexicon contains
\textit{um}, \textit{uh}, \textit{erm},
\textit{mm}, \textit{hmm}, \textit{mhm},
\textit{uhhuh}, \textit{uh-huh}, and
\textit{huh}. The restricted lexicon retains
only \textit{um}, \textit{uh}, and \textit{erm}.

An additional \texttt{pragmatic\_norm} control
applies lowercasing and colloquial expansion
while preserving fillers, backchannels, and
punctuation. This transformation leaves MRDA
unchanged; its reported neural comparisons
are single-seed supplementary controls.

Normalization implementations have corpus-specific
differences. Switchboard's original
\texttt{clean\_heavy} additionally removes
\textit{like} and \textit{youknow}, while its \texttt{gold\_alpha\_text}
uses ConvoKit's provided field, which also removes
clitics. MRDA and HVB instead use regex-based
alphanumeric filtering. Colloquial normalization
changes no Switchboard or MRDA utterances, while
punctuation removal and light cleaning leave HVB
unchanged.

As a separate single-seed sensitivity test, we
evaluate a harmonized strict heavy-normalization
condition applying
lowercasing, punctuation removal, colloquial
expansion, and deletion of only \textit{um},
\textit{uh}, and \textit{erm} across all
three corpora.

\textbf{Real-ASR Evaluation.}
We evaluate Whisper-base.en
\cite{radford-et-al2022-whisper} under two protocols.
On SLUE-HVB, models are trained and tested on
ASR transcripts (17.7\%/18.3\% train/test WER)
and compared with gold-trained/gold-tested models.
On MRDA, gold-trained models are compared on
gold and ASR test transcripts from a separately
reconstructed 60/15 meeting-disjoint split
(41.0\% WER), combining recognition errors
with train--test transcription mismatch.
Switchboard uses synthetic ASR noise because
its audio requires an LDC license.

\textbf{Evaluation.}
Macro-F1 is the primary evaluation metric,
assigning equal weight to each dialogue-act class.
Primary neural results are summarized using
mean and standard deviation across five seeds.
At the primary seed, paired Macro-F1
differences are assessed using 2,000
cluster-bootstrap resamples, with conversations
as clusters for Switchboard and calls for HVB.
McNemar's test assesses paired utterance-level
correctness rather than Macro-F1, with
Bonferroni correction across 36 comparisons; 21
remain significant at $\alpha=0.05/36$.
Because utterances within conversations are
dependent, McNemar results are supplementary
to cluster-level uncertainty estimates. Individual
$p$-values and complete condition-level outputs
are included in the public repository.

Implementation scripts and supplementary
materials are provided in a public project
repository.\footnote{
\url{https://github.com/luckymaduabuchi/asr-normalization-dialogue-acts}}

\section{Results}
\label{sec:results}

\begin{table*}[t]
\caption{Macro-F1 under original, filler-removed, and
heavily normalized transcripts. Neural results are
mean $\pm$ standard deviation over five seeds.
TF-IDF+LR is deterministic. Original corpus-specific
normalization protocols are used.}
\label{tab:grid}
\centering
\footnotesize
\setlength{\tabcolsep}{5pt}
\renewcommand{\arraystretch}{0.85}
\begin{tabular*}{\textwidth}
{@{\extracolsep{\fill}}llccc@{}}
\toprule
\textbf{Corpus} & \textbf{Model} &
\textbf{Gold F1} & \textbf{Filler F1} &
\textbf{Heavy F1} \\
\midrule
Switchboard & TF-IDF+LR & 0.311 & 0.300 & 0.293 \\
            & RoBERTa & $0.449\pm0.006$ & $0.447\pm0.005$ & $0.393\pm0.012$ \\
            & BERT    & $0.440\pm0.005$ & $0.449\pm0.008$ & $0.387\pm0.004$ \\
            & ALBERT  & $0.420\pm0.008$ & $0.409\pm0.026$ & $0.355\pm0.046$ \\
            & GPT-2   & $0.388\pm0.012$ & $0.383\pm0.009$ & $0.339\pm0.008$ \\
            & BiLSTM  & $0.370\pm0.007$ & $0.374\pm0.012$ & $0.336\pm0.010$ \\
\midrule
MRDA        & TF-IDF+LR & 0.409 & 0.364 & 0.361 \\
            & RoBERTa & $0.593\pm0.006$ & $0.550\pm0.003$ & $0.485\pm0.003$ \\
            & BERT    & $0.591\pm0.006$ & $0.556\pm0.002$ & $0.481\pm0.007$ \\
            & ALBERT  & $0.580\pm0.007$ & $0.530\pm0.016$ & $0.469\pm0.013$ \\
            & GPT-2   & $0.565\pm0.001$ & $0.522\pm0.008$ & $0.444\pm0.003$ \\
            & BiLSTM  & $0.496\pm0.005$ & $0.438\pm0.006$ & $0.401\pm0.011$ \\
\midrule
SLUE-HVB    & TF-IDF+LR & 0.713 & 0.660 & 0.661 \\
            & RoBERTa & $0.745\pm0.008$ & $0.697\pm0.004$ & $0.698\pm0.004$ \\
            & BERT    & $0.742\pm0.004$ & $0.698\pm0.004$ & $0.697\pm0.003$ \\
            & ALBERT  & $0.742\pm0.004$ & $0.697\pm0.003$ & $0.698\pm0.003$ \\
            & GPT-2   & $0.707\pm0.026$ & $0.671\pm0.004$ & $0.673\pm0.007$ \\
            & BiLSTM  & $0.732\pm0.005$ & $0.684\pm0.008$ & $0.683\pm0.011$ \\
\bottomrule
\end{tabular*}
\end{table*}

\subsection{Multi-Seed Normalization Effects}

Table~\ref{tab:grid} reports selected primary conditions
across five seeds
for the neural models and a deterministic TF-IDF+LR
baseline. Under the original preprocessing protocols,
heavy normalization reduces Macro-F1 in all 75 paired
seed--model--corpus comparisons. Filler removal also
reduces Macro-F1 in all 50 neural comparisons on MRDA
and SLUE-HVB but produces mixed effects on Switchboard.
For example, MRDA RoBERTa decreases from 0.593 to
0.550 after filler removal and to 0.485 under heavy
normalization. In contrast, Switchboard BERT improves
from 0.440 to 0.449 after filler removal but falls
to 0.387 under heavy normalization. Thus, heavy
normalization produces more consistent degradation
than filler removal alone.

The complete primary-seed nine-condition grid is
available in the repository. Across its 18
model--corpus combinations, synthetic ASR noise
degrades all 18 ($\Delta=0.033$--$0.225$).
Punctuation removal and light cleaning each degrade
11 combinations and leave seven unchanged
($\Delta=0$--$0.084$ for each operation).
Colloquial expansion and discourse-marker removal
have mixed, smaller effects ($|\Delta|\leq0.011$);
alphanumeric filtering is model-sensitive
($\Delta=-0.005$--$0.209$). The punctuation-aware
control below explains the default TF-IDF result.

\subsection{Punctuation and Filler-Lexicon Controls}

The original TF-IDF+LR baseline appeared insensitive
to punctuation removal because its default tokenizer
discarded punctuation before feature extraction.
Using a punctuation-aware, case-preserving tokenizer,
punctuation removal reduces Macro-F1 by 0.036 on
Switchboard and 0.078 on MRDA (Table~\ref{tab:punct}).
Therefore, the original near-neutral effect was a
tokenization artifact rather than an intrinsic
property of linear classifiers. SLUE-HVB remains
unaffected because its transcripts lack punctuation.

\begin{table}[t]
\caption{Punctuation-removal effects for TF-IDF+LR.
$\Delta_{\mathrm{punct}}$ denotes the Macro-F1 decrease
relative to the corresponding gold condition.}
\label{tab:punct}
\centering
\scriptsize
\setlength{\tabcolsep}{4pt}
\begin{tabular}{@{}lccc@{}}
\toprule
\textbf{Corpus} &
\textbf{Default $\Delta$} &
\textbf{Aware Gold} &
\textbf{Aware $\Delta$} \\
\midrule
Switchboard & 0.0002 & 0.360 & 0.036 \\
MRDA        & 0.0000 & 0.494 & 0.078 \\
SLUE-HVB    & 0.0000 & 0.717 & 0.000 \\
\bottomrule
\end{tabular}
\end{table}

We also evaluate a restricted filler lexicon containing
only \textit{um}, \textit{uh}, and \textit{erm}.
On MRDA, filler-removal degradation remains positive
across all six models ($\Delta=0.015$--$0.039$),
although the proportion of emptied backchannel
utterances falls from 43.1\% to 0.5\%.
Across the filtered SLUE-HVB data, the original
lexicon empties 38 of 50 backchannel utterances,
whereas the restricted lexicon empties none; only
16 backchannels occur in the held-out test set.
Accordingly, filler-removal degradation
approaches zero for the four evaluated neural models
with non-anomalous primary-seed baselines
($\Delta=-0.006$ to $0.001$).

\begin{table}[t]
\caption{Primary-seed Macro-F1 degradation under
harmonized strict heavy normalization.
$\Delta=\mathrm{gold}-\mathrm{strict}$;
positive values indicate degradation.}
\label{tab:strict}
\centering
\scriptsize
\setlength{\tabcolsep}{5pt}
\renewcommand{\arraystretch}{1.0}
\begin{tabular}{@{}lrrr@{}}
\toprule
\textbf{Model} &
\textbf{SWB} & \textbf{MRDA} & \textbf{HVB} \\
\midrule
TF-IDF+LR & 0.007 & 0.039 & 0.015 \\
RoBERTa   & 0.044 & 0.080 & $-0.004$ \\
BERT      & 0.041 & 0.076 & $-0.001$ \\
ALBERT    & 0.032 & 0.077 & 0.003 \\
GPT-2     & 0.048 & 0.110 & $-0.049$ \\
BiLSTM    & 0.022 & 0.056 & 0.000 \\
\bottomrule
\end{tabular}
\end{table}

Table~\ref{tab:strict} reports results under an
identical strict heavy-normalization function.
All six models exhibit Macro-F1 degradation on
Switchboard and MRDA, whereas five of six SLUE-HVB models change
by at most 0.015. The exception is GPT-2, which
improves by 0.049 from an unusually low gold
baseline. All strict-control results use the primary seed.

Separately, omitting Switchboard's extra
\textit{like}/\textit{youknow} removal while retaining the
nine-word lexicon changes the normalized text
for 17.7\% of utterances. TF-IDF+LR Macro-F1
changes from 0.2933 to 0.2954; corresponding
neural reruns are unavailable.

At the primary seed, \texttt{pragmatic\_norm} Macro-F1
remains within $\pm0.013$ of gold across 12
Switchboard and SLUE-HVB model combinations.
It outperforms heavy normalization in 11 of 12
comparisons, suggesting that mild standardization
may preserve performance when discourse cues remain.

\subsection{Statistical and Class-Level Analysis}

Five-seed standard deviations summarize training
variability. Paired cluster-bootstrap confidence
intervals quantify test-cluster sampling uncertainty
at primary seed, not uncertainty in five-seed means.
Throughout, positive $\Delta$ denotes the
Macro-F1 decrease relative to gold.

At the primary seed, cluster-bootstrap 95\% confidence
intervals for heavy-normalization Macro-F1 differences
exclude zero for all five neural models on Switchboard
and four of five on SLUE-HVB. Filler-removal intervals
exclude zero for only ALBERT on Switchboard and four
neural models on SLUE-HVB. For TF-IDF+LR on Switchboard,
filler-removal and heavy-normalization differences are
0.011 $[0.003,0.020]$ and 0.017 $[0.009,0.027]$,
respectively. These are nominal 95\% intervals without
multiplicity adjustment; an F1 decrease alone does
not establish statistical significance. The reformatted
primary MRDA records lack meeting identifiers. On a
separate exploratory 60/15 meeting-disjoint reconstruction, heavy-
normalization intervals exclude zero for four of five
neural models, while filler-only effects are split-sensitive.

Utterance-level McNemar tests provide an additional
paired-correctness comparison but are not tests
of Macro-F1 and do not adjust for conversation
clustering; cluster-bootstrap intervals therefore
remain the primary uncertainty evidence.

\begin{table}[t]
\caption{Selected class-level effects of lexicon-based
removal for deterministic TF-IDF+LR on held-out test
splits. P, R, and F1 denote precision, recall, and
F1-score. BC denotes SLUE-HVB backchannels.}
\label{tab:class-effects}
\centering
\scriptsize
\setlength{\tabcolsep}{4pt}
\renewcommand{\arraystretch}{1.05}
\begin{tabular}{@{}llcrr@{}}
\toprule
\textbf{Corpus} & \textbf{Class} &
\textbf{Metric} & \textbf{Gold} & \textbf{No Fill.} \\
\midrule
Switchboard & b  & P  & 0.558 & 0.436 \\
            & b  & F1 & 0.433 & 0.365 \\
\midrule
MRDA        & fh & R  & 0.495 & 0.330 \\
            & h  & R  & 0.313 & 0.104 \\
\midrule
SLUE-HVB    & BC & P  & 0.353 & 0.092 \\
            & BC & R  & 0.750 & 0.750 \\
            & BC & F1 & 0.480 & 0.163 \\
\bottomrule
\end{tabular}
\end{table}

Table~\ref{tab:class-effects} shows different
class-level effects across corpora. On Switchboard,
backchannel precision decreases from 0.558 to 0.436,
with an F1 reduction of 0.068. On MRDA, recall
decreases for floor holders (\texttt{fh}) and holds
before answers or agreements (\texttt{h}),
indicating difficulty recognizing turn-management
dialogue acts. In contrast, SLUE-HVB backchannel
recall remains at 0.750, while precision decreases
from 0.353 to 0.092, reducing F1 from 0.480 to
0.163. Because Macro-F1 weights this low-support class
equally with the other 15 classes, its large F1
change warrants cautious interpretation.
The restricted-lexicon analysis indicates that
much of this degradation results from deleting
class-defining backchannel expressions.

\subsection{Real-ASR Evaluation}

Whisper-base.en is evaluated under two protocols.
On SLUE-HVB, models are trained and tested on
Whisper transcripts (17.7\%/18.3\% train/test WER).
Five models exhibit lower Macro-F1, whereas
GPT-2 improves from 0.6674 to 0.6765 at primary seed.
This isolated result does not establish an ASR
benefit, especially given its unusually low baseline.

On MRDA, models trained on gold transcripts are
evaluated on Whisper transcripts from 15 held-out
meetings (WER: 41.0\%). All six models degrade,
with Macro-F1 decreases ranging from 0.1366 to
0.2822. Unlike SLUE-HVB, this experiment combines
recognition errors with train--test transcription
mismatch; the effect sizes are therefore not directly
comparable.

\section{Discussion and Conclusion}
\label{sec:conclusion}

Our findings show that transcript normalization affects
dialogue act classification differently across models,
corpora, and preprocessing operations. Under the original
protocols, heavy normalization reduces Macro-F1 across
all five seeds for every neural model--corpus combination,
whereas filler removal produces mixed effects on
Switchboard but consistent degradation on MRDA and
SLUE-HVB.

\textbf{Interpretation of normalization effects.}
The punctuation-aware TF-IDF control shows
that punctuation is predictive when the feature
extractor preserves it. The default TF-IDF
baseline appeared insensitive because its tokenizer
discarded punctuation. Filler-lexicon controls
reveal a different mechanism: on MRDA, degradation
persists when removal is restricted to \textit{um},
\textit{uh}, and \textit{erm}, suggesting that these
expressions provide useful classification cues.
On SLUE-HVB, most degradation disappears under the
restricted lexicon because the original inventory
removes class-defining backchannel expressions.
Thus, not all performance losses reflect the same
type of information removal.

\textbf{Robustness and practical implications.}
\label{sec:swb-recovery}
The five-seed experiments show that heavy-normalization
degradation is more consistent than filler-removal
effects. Under harmonized strict normalization,
degradation persists across all six classifiers on
Switchboard and MRDA but is largely absent on
SLUE-HVB, demonstrating sensitivity to corpus
characteristics and lexicon definitions.
Single-seed pragmatic-normalization controls
further suggest that modest text standardization
can retain dialogue-act performance when
backchannels and punctuation are preserved.

\textbf{Limitations.}
The study is limited to three English-language
corpora. SLUE-HVB findings concern a first-tag,
single-label reduction and do not evaluate its full
multi-label dialogue-act task. The original Switchboard heavy-normalization
implementation additionally removes \textit{like} and \textit{youknow},
and harmonized lexicon experiments use a single seed.
On Switchboard, spaced backchannel expressions such
as \textit{Uh - huh} are partially affected by the
strict lexicon, which removes \textit{uh} but retains
\textit{huh}. Thus, identical normalization rules
do not necessarily remove equivalent linguistic cues
across corpora. According to the release-defined
canonical split, the primary MRDA partition is
meeting-disjoint but not speaker-disjoint; because
meeting identifiers are absent from the reformatted records,
meeting-level confidence intervals are unavailable
for that primary split. The alternative MRDA
neural comparisons use different training scripts,
hyperparameters, and text formatting, preventing
controlled attribution of their differences to
normalization. Real-ASR evaluations use different
protocols: matched ASR training/testing on SLUE-HVB
and gold-trained/ASR-tested models on MRDA.
The pragmatics-preserving control uses reference
transcripts and does not recover cues already
missing from real ASR output. All reported BiLSTM
results use the corrected empty-input pooling
implementation.

Overall, aggressive transcript normalization can
degrade dialogue act classification, but the
magnitude and mechanism depend on feature extraction,
lexicon design, and conversational setting.
These findings support selective, task-dependent
normalization rather than indiscriminate transcript
cleaning. Future work will examine multi-seed harmonized
normalization, matched ASR training on MRDA, and
additional conversational domains.

\bibliographystyle{IEEEtran}
\bibliography{references}

@article{CLARK1989259,
title = {Contributing to discourse},
journal = {Cognitive Science},
volume = {13},
number = {2},
pages = {259-294},
year = {1989},
issn = {0364-0213},
doi = {https://doi.org/10.1016/0364-0213(89)90008-6},
url = {https://www.sciencedirect.com/science/article/pii/0364021389900086},
author = {Herbert H. Clark and Edward F. Schaefer}
}

@article{CLARK200273,
title = {Using uh and um in spontaneous speaking},
journal = {Cognition},
volume = {84},
number = {1},
pages = {73-111},
year = {2002},
issn = {0010-0277},
doi = {https://doi.org/10.1016/S0010-0277(02)00017-3},
url = {https://www.sciencedirect.com/science/article/pii/S0010027702000173},
author = {Herbert H. Clark and Jean E. {Fox Tree}}
}

@article{https://doi.org/10.1111/j.1749-818X.2008.00068.x,
author = {Corley, Martin and Stewart, Oliver W.},
title = {Hesitation Disfluencies in Spontaneous Speech: The Meaning of um},
journal = {Language and Linguistics Compass},
volume = {2},
number = {4},
pages = {589-602},
doi = {https://doi.org/10.1111/j.1749-818X.2008.00068.x},
url = {https://compass.onlinelibrary.wiley.com/doi/abs/10.1111/j.1749-818X.2008.00068.x},
eprint = {https://compass.onlinelibrary.wiley.com/doi/pdf/10.1111/j.1749-818X.2008.00068.x},
year = {2008}
}

@techreport{jurafsky-shriberg-biasca1997,
  title = {Switchboard {SWBD-DAMSL} Shallow-Discourse-Function Annotation Coders Manual, Draft 13},
  author = {Jurafsky, Daniel and Shriberg, Elizabeth and Biasca, Debra},
  institution = {University of Colorado, Boulder Institute of Cognitive Science},
  year = {1997},
  number = {97-02},
  url = {https://web.stanford.edu/~jurafsky/ws97/manual.august1.html}
}

@inproceedings{Core1997CodingDW,
  title={Coding Dialogs with the DAMSL Annotation Scheme},
  author={Mark G. Core and James F. Allen},
  year={1997},
  url={https://api.semanticscholar.org/CorpusID:18744825}
}

@inproceedings{chang-et-al2020,
  title = {ConvoKit: A Toolkit for the Analysis of Conversations},
  author = {Chang, Jonathan P. and Chiam, Caleb and Fu, Liye and Wang, Andrew and Zhang, Justine and Danescu-Niculescu-Mizil, Cristian},
  booktitle = {Proceedings of the 21st Annual Meeting of the Special Interest Group on Discourse and Dialogue},
  year = {2020},
  address = {1st virtual meeting},
  publisher = {Association for Computational Linguistics},
  pages = {57--60},
  doi = {10.18653/v1/2020.sigdial-1.8},
  url = {https://aclanthology.org/2020.sigdial-1.8/}
}

@inproceedings{shriberg1999phoneticCO,
  title={Phonetic Consequences of Speech Disfluency},
  author={Elizabeth Shriberg},
  year={1999},
  url={https://api.semanticscholar.org/CorpusID:14273237}
}

@article{Cui2021AnAT,
  title={An Approach to Improve Robustness of NLP Systems against ASR Errors},
  author={Tong Cui and Jinghui Xiao and Liangyou Li and Xin Jiang and Qun Liu},
  journal={ArXiv},
  year={2021},
  volume={abs/2103.13610},
  url={https://api.semanticscholar.org/CorpusID:232352551}
}

@Article{Nechaev2021,
 author = {Yaroslav Nechaev and Weitong Ruan and Imre Kiss},
 title = {Towards NLU model robustness to ASR errors at scale},
 year = {2021},
 url = {https://www.amazon.science/publications/towards-nlu-model-robustness-to-asr-errors-at-scale},
}

@article{LI2024104598,
title = {Useful blunders: Can automated speech recognition errors improve downstream dementia classification?},
journal = {Journal of Biomedical Informatics},
volume = {150},
pages = {104598},
year = {2024},
issn = {1532-0464},
doi = {https://doi.org/10.1016/j.jbi.2024.104598},
url = {https://www.sciencedirect.com/science/article/pii/S1532046424000169},
author = {Changye Li and Weizhe Xu and Trevor Cohen and Serguei Pakhomov}
}

@inproceedings{kubis-etal-2023-back,
    title = "Back Transcription as a Method for Evaluating Robustness of Natural Language Understanding Models to Speech Recognition Errors",
    author = "Kubis, Marek  and
      Sk{\'o}rzewski, Pawe{\l}  and
      Sowa{\'n}ski, Marcin  and
      Zi{\k{e}}tkiewicz, Tomasz",
    editor = "Bouamor, Houda  and
      Pino, Juan  and
      Bali, Kalika",
    booktitle = "Proceedings of the 2023 Conference on Empirical Methods in Natural Language Processing",
    month = dec,
    year = "2023",
    address = "Singapore",
    publisher = "Association for Computational Linguistics",
    url = "https://aclanthology.org/2023.emnlp-main.724/",
    doi = "10.18653/v1/2023.emnlp-main.724",
    pages = "11824--11835"
}

@article{GinzburgFernandezSchlangen2014,
	Author = {Ginzburg, Jonathan and Fern\'andez, Raquel and David, Schlangen},
	Journal = {Semantics and Pragmatics},
	Volume = {7},
	Number = {9},
	Pages = {1--64},
	Title = {Disfluencies as intra-utterance dialogue moves},
  url = {https://semprag.org/index.php/sp/article/view/sp.7.9/pdfsp79},
	doi = {http://dx.doi.org/10.3765/sp.7.9},
	Year = {2014}
}

@inproceedings{zusag24_interspeech,
  title     = {{CrisperWhisper: Accurate Timestamps on Verbatim Speech Transcriptions}},
  author    = {Mario Zusag and Laurin Wagner and Bernhad Thallinger},
  year      = {2024},
  booktitle = {{Interspeech 2024}},
  pages     = {1265--1269},
  doi       = {10.21437/Interspeech.2024-731},
  issn      = {2958-1796},
}

@article{stolcke-etal-2000-dialogue,
    title = "Dialogue act modeling for automatic tagging and recognition of conversational speech",
    author = "Stolcke, Andreas  and
      Ries, Klaus  and
      Coccaro, Noah  and
      Shriberg, Elizabeth  and
      Bates, Rebecca  and
      Jurafsky, Daniel  and
      Taylor, Paul  and
      Martin, Rachel  and
      Van Ess-Dykema, Carol  and
      Meteer, Marie",
    journal = "Computational Linguistics",
    volume = "26",
    number = "3",
    year = "2000",
    address = "Cambridge, MA",
    publisher = "MIT Press",
    url = "https://aclanthology.org/J00-3003/",
    pages = "339--374"
}

@inproceedings{shapira-etal-2025-measuring,
    title = "Measuring the Effect of Transcription Noise on Downstream Language Understanding Tasks",
    author = "Shapira, Ori  and
      Chazan, Shlomo E.  and
      Cohen, Amir DN",
    editor = "Che, Wanxiang  and
      Nabende, Joyce  and
      Shutova, Ekaterina  and
      Pilehvar, Mohammad Taher",
    booktitle = "Proceedings of the 63rd Annual Meeting of the Association for Computational Linguistics (Volume 1: Long Papers)",
    month = jul,
    year = "2025",
    address = "Vienna, Austria",
    publisher = "Association for Computational Linguistics",
    url = "https://aclanthology.org/2025.acl-long.1449/",
    doi = "10.18653/v1/2025.acl-long.1449",
    pages = "29978--30004",
    ISBN = "979-8-89176-251-0"
}

@article{zelasko-etal-2021-helps,
    title = "What Helps Transformers Recognize Conversational Structure? Importance of Context, Punctuation, and Labels in Dialog Act Recognition",
    author = "{\.Z}elasko, Piotr  and
      Pappagari, Raghavendra  and
      Dehak, Najim",
    editor = "Roark, Brian  and
      Nenkova, Ani",
    journal = "Transactions of the Association for Computational Linguistics",
    volume = "9",
    year = "2021",
    address = "Cambridge, MA",
    publisher = "MIT Press",
    url = "https://aclanthology.org/2021.tacl-1.69/",
    doi = "10.1162/tacl_a_00420",
    pages = "1163--1179"
}

@inproceedings{shriberg-et-al2004-mrda,
  title = {The {ICSI} Meeting Recorder Dialog Act ({MRDA}) Corpus},
  author = {Shriberg, Elizabeth and Dhillon, Raj and Bhagat, Sonali and Ang, Jeremy and Carvey, Hannah},
  booktitle = {Proceedings of the 5th {SIGdial} Workshop on Discourse and Dialogue at {HLT-NAACL} 2004},
  year = {2004},
  address = {Cambridge, Massachusetts, USA},
  publisher = {Association for Computational Linguistics},
  pages = {97--100},
  url = {https://aclanthology.org/W04-2319/}
}

@misc{wylupek-mrda,
  title = {{MRDA} Corpus ({H}ugging{F}ace reformatting of the {ICSI} {MRDA} Corpus)},
  author = {Wylupek},
  year = {2025},
  howpublished = {\url{https://huggingface.co/datasets/wylupek/mrda-corpus}},
  note = {Reformatted release of \cite{shriberg-et-al2004-mrda} for use with the HuggingFace \texttt{datasets} library; accessed July 2026}
}

@article{devlin-et-al2018-bert,
  title = {{BERT}: Pre-training of Deep Bidirectional Transformers for Language Understanding},
  author = {Devlin, Jacob and Chang, Ming-Wei and Lee, Kenton and Toutanova, Kristina},
  year = {2018},
  howpublished = {arXiv preprint arXiv:1810.04805},
  url = {https://arxiv.org/abs/1810.04805}
}

@article{liu-et-al2019-roberta,
  title = {{RoBERTa}: A Robustly Optimized {BERT} Pretraining Approach},
  author = {Liu, Yinhan and Ott, Myle and Goyal, Naman and Du, Jingfei and Joshi, Mandar and Chen, Danqi and Levy, Omer and Lewis, Mike and Zettlemoyer, Luke and Stoyanov, Veselin},
  year = {2019},
  howpublished = {arXiv preprint arXiv:1907.11692},
  url = {https://arxiv.org/abs/1907.11692}
}

@article{lan-et-al2019-albert,
  title = {{ALBERT}: A Lite {BERT} for Self-supervised Learning of Language Representations},
  author = {Lan, Zhenzhong and Chen, Mingda and Goodman, Sebastian and Gimpel, Kevin and Sharma, Piyush and Soricut, Radu},
  year = {2019},
  howpublished = {arXiv preprint arXiv:1909.11942},
  url = {https://arxiv.org/abs/1909.11942}
}

@misc{radford-et-al2019-gpt2,
  title = {Language Models are Unsupervised Multitask Learners},
  author = {Radford, Alec and Wu, Jeffrey and Child, Rewon and Luan, David and Amodei, Dario and Sutskever, Ilya},
  year = {2019},
  howpublished = {OpenAI Blog, vol.\ 1, no.\ 8}
}

@inproceedings{shon-et-al2023-slue2,
  title = {{SLUE} Phase-2: A Benchmark Suite of Diverse Spoken Language Understanding Tasks},
  author = {Shon, Suwon and Arora, Siddhant and Lin, Chyi-Jiunn and Pasad, Ankita and Wu, Felix and Sharma, Roshan and Wu, Wei-Lun and Lee, Hung-yi and Livescu, Karen and Watanabe, Shinji},
  booktitle = {Proceedings of the 61st Annual Meeting of the Association for Computational Linguistics (Volume 1: Long Papers)},
  year = {2023},
  pages = {8906--8937},
  publisher = {Association for Computational Linguistics},
  doi = {10.18653/v1/2023.acl-long.496},
  url = {https://aclanthology.org/2023.acl-long.496/}
}

@article{hochreiter-schmidhuber1997-lstm,
  title = {Long Short-Term Memory},
  author = {Hochreiter, Sepp and Schmidhuber, J{\"u}rgen},
  journal = {Neural Computation},
  year = {1997},
  volume = {9},
  number = {8},
  pages = {1735--1780},
  doi = {10.1162/neco.1997.9.8.1735}
}

@inproceedings{radford-et-al2022-whisper,
  title = {Robust Speech Recognition via Large-Scale Weak Supervision},
  author = {Radford, Alec and Kim, Jong Wook and Xu, Tao and Brockman, Greg and McLeavey, Christine and Sutskever, Ilya},
  booktitle = {Proceedings of the 40th International Conference on Machine Learning},
  year = {2023},
  url = {https://arxiv.org/abs/2212.04356}
}

@inproceedings{onyekwelu2025emotion,
  title={Emotion detection in speech using lightweight and transformer-based models: A comparative and ablation study},
  author={Onyekwelu-Udoka, Lucky and Islam, Md Shafiqul and Hasan, Md Shahedul},
  booktitle={2025 IEEE 2nd International Conference on Computing, Applications and Systems (COMPAS)},
  pages={1--6},
  year={2025},
  organization={IEEE}
}

\end{document}